\documentclass[conference]{IEEEtran}
\IEEEoverridecommandlockouts
\usepackage{cite}
\usepackage{amsmath,amssymb,amsfonts}
\usepackage{algorithmic}
\usepackage{graphicx}
\usepackage{textcomp}
\usepackage{xcolor}
\usepackage{url}
\usepackage{booktabs}
\usepackage{array}
\def\BibTeX{{\rm B\kern-.05em{\sc i\kern-.025em b}\kern-.08em
    T\kern-.1667em\lower.7ex\hbox{E}\kern-.125emX}}
\begin{document}

\title{Parking Assistance for Trailer-Truck Transport Vehicles Using Sensor Fusion and Motion Planning}

\author{
\IEEEauthorblockN{
George Alenchery, Tova Quinones, Thomas Jeske, Tristan Lindo-Slones,\\
Amber Jones, Jordan Fletcher, Lentz Fortune
}
\IEEEauthorblockA{
\textit{Software Engineering Department}\\
\textit{Florida Gulf Coast University}\\
\{gaalenchery6113, tdquinones135, trjeske5899, talindoslones4387,\\
anjones4463, jbfletcher0803, lgfortune9567\}@eagle.fgcu.edu
}
\and
\IEEEauthorblockN{Sudip Dhakal}
\IEEEauthorblockA{
\textit{Software Engineering Department}\\
\textit{Florida Gulf Coast University}\\
sdhakal@fgcu.edu
}
}

\maketitle

\begin{abstract}
Autonomous driving technology has rapidly evolved over the past decade, offering
significant improvements in transportation efficiency, safety, and cost reduction.
While much of the progress has focused on highway driving and obstacle avoidance,
low-speed maneuvers such as parking remain among the most difficult challenges for
autonomous systems. This challenge is especially pronounced in trailer-truck
transport vehicles due to their articulated motion and environmental constraints.
This paper presents a proposed framework for autonomous truck parking that integrates
perception, motion planning, control systems, and infrastructure awareness. By
combining sensor fusion, Hybrid A* path planning, nonlinear model predictive control
(NMPC), and data-driven parking systems, this work highlights the importance of
system-level coordination for reliable and scalable autonomous parking solutions.
As a proof-of-concept implementation, we adapted an open-source A* path planning
simulation to incorporate a tractor-trailer kinematic model, demonstrating
articulated vehicle path planning within a command-line simulation environment,
with jackknife prevention identified as an area requiring further development.
\end{abstract}



\section{Introduction}

Autonomous driving technology has rapidly evolved over the past decade, offering
significant improvements in transportation efficiency, safety, and cost reduction.
While much of the progress has focused on highway driving and obstacle avoidance,
low-speed maneuvers such as parking remain among the most difficult challenges for
autonomous systems.

This challenge is especially pronounced in trailer-truck transport vehicles, where
the complexity of articulated motion and environmental constraints introduces
significant difficulties. Unlike standard passenger vehicles, semi-trucks operate
as articulated systems composed of a tractor and a trailer connected by a hitch.
This structure creates nonlinear dynamics that make reversing and parking far more
difficult than forward motion. The trailer introduces additional degrees of freedom
that must be actively managed, and failure to do so risks dangerous conditions such
as jackknifing or trailer swing.

Beyond vehicle control challenges, the problem extends to infrastructure limitations.
Safe and accessible parking spaces for large vehicles are scarce, particularly in
urban environments. As automation increases within the trucking industry, the need
for intelligent, infrastructure-aware parking systems becomes critical.

This paper makes two distinct contributions. First, it proposes a unified system
architecture for autonomous truck parking by synthesizing existing research across
perception, planning, control, and infrastructure awareness into a single cohesive
design. This proposed architecture was not fully implemented; it represents a
design target informed by the literature review. Second, we describe a
proof-of-concept simulation in which an open-source A* path planning codebase was
adapted to include a truck-trailer kinematic model, providing an initial
implementation baseline for future development. The remainder of this paper is
organized as follows: Section~\ref{sec:background} provides relevant background,
Section~\ref{sec:litreview} surveys the related literature,
Section~\ref{sec:method} presents the proposed system architecture,
Section~\ref{sec:results} describes our implementation and results,
Section~\ref{sec:discussion} discusses findings and challenges, and
Section~\ref{sec:conclusion} concludes the paper.


\section{Background}
\label{sec:background}

Autonomous driving systems are typically divided into several functional layers:
sensing, perception, localization, planning, and control. Each layer contributes
to safe and reliable vehicle operation, and a failure at any layer can compromise
the entire system.

Parking differs fundamentally from general driving tasks because it prioritizes
precision over speed. Vehicles must navigate tight spaces, avoid static and dynamic
obstacles, and align accurately with designated parking areas. These requirements
demand higher positional accuracy than highway or urban driving scenarios.

For trucks, these challenges are compounded by large vehicle dimensions and
articulated motion dynamics. The tractor-trailer configuration behaves differently
depending on whether the vehicle is moving forward or reversing. During reversing
maneuvers, the trailer responds inversely to steering inputs, making path following
significantly more difficult. Additionally, the extended length of the combined
vehicle increases the swept area during turns, raising the risk of collision with
infrastructure or other vehicles. Understanding these dynamics is essential context
for the system design decisions discussed in subsequent sections.


\section{Literature Review}
\label{sec:litreview}

The development of an autonomous truck parking system requires integration across
multiple research domains, each contributing a critical layer to the overall
architecture. This section synthesizes the key findings from the reviewed literature
into a unified system-level perspective.

Motion planning forms the foundation of any autonomous parking solution. Hybrid A*
path planning is particularly well-suited for articulated vehicles because it
incorporates nonholonomic vehicle kinematics directly into the search process,
producing trajectories that are both collision-free and physically realizable
\cite{hybrid}. This is complemented by Nonlinear Model Predictive Control (NMPC),
which refines trajectories in real time by minimizing a cost function subject to
constraints such as obstacle avoidance, steering limits, and hitch angle bounds
\cite{mpc}. A Linear Quadratic Regulator (LQR) can then be applied as a tracking
layer to minimize deviation from the planned path with computational efficiency.

Reliable perception underpins safe parking execution. Transformer-based sensor
fusion architectures, such as TransFuser, extend beyond local geometric alignment
by enabling global attention across camera and LiDAR inputs \cite{transfuser}.
This allows the system to reason about spatially distributed features, such as
parking boundaries and dynamic obstacles, within a unified environmental
representation. Bird's Eye View (BEV) projections further simplify downstream
planning by providing a structured, sensor-agnostic top-down scene description.

Infrastructure awareness is equally essential, as the availability and legality
of parking locations directly constrain the planning problem. Data-driven
classification frameworks leveraging GPS trajectories and geospatial datasets
enable systematic identification of viable truck parking areas, evaluated along
rdimensions of accessibility, legality, and roadside status \cite{parkingdata}.
Shared mapping systems extend this capability by aggregating real-time contributions
from multiple vehicles, enabling dynamic updates to parking availability.

Deep reinforcement learning (DRL) offers a complementary, policy-based perspective
on parking \cite{drl}. By learning control strategies directly from environmental
interaction using BEV state representations, DRL methods can produce adaptive
behaviors without explicit trajectory computation. However, their high training
demands and limited interpretability suggest they are best deployed in concert with
classical planners rather than as standalone solutions.

Lastly, through the coordination of various agents, the foundations for future
scalable deployment are laid out. Reservation-based architectures allow vehicles to
negotiate and secure parking spots in advance, while dynamic conflict resolution
mechanisms reduce congestion in shared environments.

Taken together, the reviewed literature points toward a layered architecture
comprising sensor fusion, infrastructure modeling, Hybrid A* or NMPC-based
trajectory planning, LQR control, and multi-agent coordination as the most
promising framework for autonomous truck parking. Our work builds directly on
this foundation: we adopt Hybrid A* as the core planning algorithm based on its
demonstrated suitability for nonholonomic articulated vehicles, and we use the
infrastructure classification framework of \textit{"A data driven systematic
approach for identifying and classifying long haul truck parking
locations"} \cite{parkingdata} as the basis for
our proposed parking space identification layer. Where the reviewed systems
primarily address passenger vehicle parking or single-body vehicle dynamics, our
contribution specifically targets the tractor-trailer kinematic coupling that
makes truck parking a qualitatively harder problem. Table~\ref{tab:comparison}
in Section~\ref{sec:method} summarizes how our approach compares to the reviewed
methods at a feature level.


\section{Proposed System Architecture}
\label{sec:method}

It is important to note that the architecture described in this section is a
\textit{proposed design} synthesized from the reviewed literature. It represents
the full system that would be required for production-ready autonomous truck
parking. The components that were actually implemented as part of this project
are described separately in Section~\ref{sec:results}. The proposed architecture
is included here to situate our proof-of-concept within a complete system context
and to serve as a roadmap for future development.

\begin{figure}[htbp]
\centering
\includegraphics[width=\columnwidth]{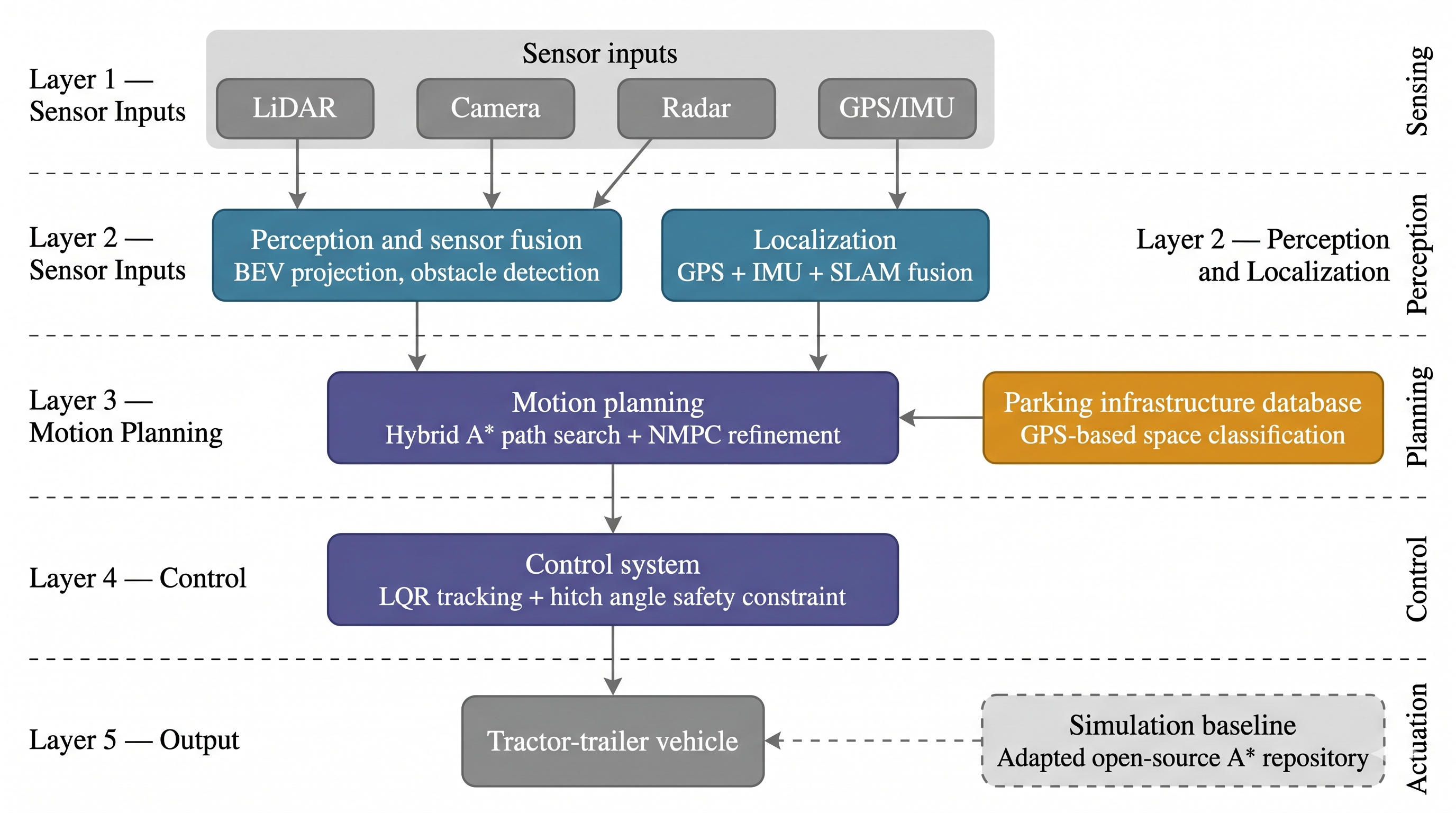}
\caption{Proposed layered architecture for autonomous truck parking. Sensor inputs
flow through perception and localization into motion planning, which queries a
parking infrastructure database before generating a trajectory. The control system
executes the trajectory subject to a hitch-angle safety constraint. Note that this
is a proposed design; only the path planning and kinematic modeling components
were implemented in this work.}
\label{fig:arch}
\end{figure}

The proposed system is organized as a layered pipeline in which each stage
processes the output of the stage above it and passes its result downstream.
Figure~\ref{fig:arch} illustrates the full architecture. At the top, raw sensor
data from LiDAR, cameras, radar, and GPS/IMU units are fused into a unified
environmental model. The perception layer produces a Bird's Eye View scene
representation, while the localization layer maintains a continuous pose estimate
of both the tractor and the trailer. These outputs feed into the motion planning
stage, which queries a parking infrastructure database to identify candidate spaces
and then generates a collision-free trajectory using Hybrid A* search. A secondary
NMPC layer refines that trajectory before it is passed to the control system, which
issues low-level steering, throttle, and braking commands to the vehicle. A
hitch-angle safety constraint is enforced at the control layer at all times to
prevent jackknifing.

The following subsections describe each proposed layer and the design rationale
behind the algorithmic choices.

\subsection{Perception and Sensor Fusion}

Accurate perception is fundamental to autonomous parking. Vehicles rely on multiple
sensors, including cameras, LiDAR, radar, and GPS, each of which provides
complementary information about the environment.

Cameras capture rich semantic details such as lane markings, signage, and
pedestrians, but lack reliable depth estimation. LiDAR provides precise 3D spatial
point clouds but does not capture semantic context. Radar offers robust performance
in adverse weather but at lower spatial resolution. Combining these modalities
through sensor fusion produces a more complete and reliable environmental model
than any single sensor alone \cite{sensor}.

The proposed architecture uses a transformer-based fusion approach, motivated by
architectures such as TransFuser \cite{transfuser}, which enables global attention
across all sensor inputs simultaneously. Fused sensor data is projected into a
Bird's Eye View (BEV) representation, which provides a structured, top-down
perspective of the environment and simplifies downstream planning by decoupling
spatial layout from sensor-specific viewing angles. This component was not
implemented in the current work.

\subsection{Localization}

Localization determines the vehicle's precise position and orientation within
its environment. For parking, localization must be accurate to within centimeters,
as small positional errors can result in misalignment with the parking target or
collision with nearby obstacles.

GPS alone is insufficient for this level of accuracy, particularly in urban
environments where signal multipath and occlusion from buildings degrade
positioning quality. The proposed architecture fuses GPS with inertial measurement
units (IMUs), wheel odometry, and LiDAR-based simultaneous localization and
mapping (SLAM) techniques. For truck parking specifically, localization must also
account for the relative pose between the tractor and trailer, as the hitch angle
directly affects trajectory planning. This component was not implemented in the
current work.

\subsection{Motion Planning}
\label{sec:planning}

Motion planning generates a feasible, collision-free path from the vehicle's
current state to the desired parking position. For articulated vehicles, this
involves satisfying complex kinematic and dynamic constraints simultaneously.

The proposed architecture uses Hybrid A* as the primary planning algorithm because
it extends the classical A* graph search with a continuous vehicle state
representation \cite{hybrid}. By incorporating the vehicle's steering and motion
model directly into the search, Hybrid A* guarantees that generated paths respect
the nonholonomic constraints of the tractor-trailer system, including minimum
turning radius and maximum hitch angle.

To refine and optimize these trajectories, NMPC is applied as a secondary
planning layer. NMPC solves the following optimization problem at each time step:

\begin{equation}
\min_{u} \sum_{k=0}^{N} \left( (x_{ref,k} - x_k)^2 + \lambda u_k^2 \right)
\end{equation}

where $x_k$ is the vehicle state, $x_{ref,k}$ is the reference trajectory,
$u_k$ is the control input, and $\lambda$ is a regularization weight. This
formulation penalizes both tracking error and excessive control effort,
producing smooth and feasible parking trajectories \cite{mpc}. The NMPC layer
was not implemented in the current work; path planning in our proof-of-concept
uses A* search with B-spline smoothing as provided by the baseline repository.

\subsection{Control System}

The proposed control system executes the planned trajectory by generating low-level
actuator commands for steering, throttle, and braking. We propose using a Linear
Quadratic Regulator (LQR) as the trajectory tracking controller due to its
computational efficiency and well-understood stability properties. The LQR
minimizes the following cost functional:

\begin{equation}
J = \sum_{k=0}^{\infty} \left( x_k^T Q x_k + u_k^T R u_k \right)
\end{equation}

where $Q$ and $R$ are positive semi-definite weighting matrices that balance
state deviation against control effort. A critical safety constraint is the hitch
angle limit: if the angle between the tractor and trailer exceeds a threshold
during reversing, the vehicle enters a jackknifing condition. The control system
must enforce a hard constraint on the hitch angle at all times. In our
proof-of-concept, trajectory control is handled by the MPC controller provided in
the baseline repository rather than a separately implemented LQR layer.

\subsection{Parking Infrastructure and Data Systems}
\label{sec:infra}

The proposed architecture includes a data-driven infrastructure layer that builds
spatial databases of parking locations from GPS trajectory data and geospatial
datasets \cite{parkingdata}. Each candidate location is evaluated against
attributes such as physical dimensions, surface type, legal status, proximity to
roadways, and accessibility for large vehicles. This component was not implemented
in the current work; parking spot selection in our simulation is handled by a
fixed 24-spot grid environment.

\subsection{Comparison with Related Work}

Table~\ref{tab:comparison} summarizes how the proposed architecture compares to
the key systems reviewed in the literature across five feature dimensions.

\begin{table}[htbp]
\caption{Feature Comparison: Proposed Architecture vs. Related Work}
\label{tab:comparison}
\centering
\begin{tabular}{>{\raggedright\arraybackslash}p{2.2cm}
                >{\raggedright\arraybackslash}p{0.85cm}
                >{\raggedright\arraybackslash}p{0.85cm}
                >{\raggedright\arraybackslash}p{0.85cm}
                >{\raggedright\arraybackslash}p{0.85cm}
                >{\raggedright\arraybackslash}p{0.85cm}}
\toprule
\textbf{Feature} &
\textbf{Hybrid A* \cite{hybrid}} &
\textbf{Trans- Fuser \cite{transfuser}} &
\textbf{DRL \cite{drl}} &
\textbf{Infra. \cite{parkingdata}} &
\textbf{Ours} \\
\midrule
Articulated vehicle support   & Yes  & No   & No   & No   & Yes \\
Sensor fusion                 & No   & Yes  & No   & No   & Proposed \\
Infrastructure awareness      & No   & No   & No   & Yes  & Proposed \\
Real-time trajectory refine.  & No   & No   & No   & No   & Proposed \\
Trailer kinematic model       & No   & No   & No   & No   & Yes \\
Simulation validated          & Yes  & Yes  & Yes  & No   & Partial \\
\bottomrule
\end{tabular}
\end{table}


\section{Implementation and Results}
\label{sec:results}

This section describes what was actually built and tested as part of this project,
distinct from the proposed architecture in Section~\ref{sec:method}.

\subsection{Simulation Environment Selection}

Our implementation process went through several environment evaluations before
arriving at a working baseline. We initially planned to develop the system within
CARLA, a high-fidelity autonomous driving simulator. However, CARLA's large
installation footprint made it impractical to run on team members' personal
machines, and scheduling conflicts limited consistent access to the university
lab environment where it was available. This made iterative development within
CARLA unsustainable for our project timeline.

We subsequently considered Gazebo and MATLAB as alternatives. Gazebo was
unfamiliar to the team and would have required significant ramp-up time before
any truck-specific work could begin. MATLAB, while familiar in other contexts
such as linear algebra, was not an area of experience for the team in the context
of visual simulation or vehicle modeling, and again, consistent lab access was
a limiting factor.

Given these constraints, we pivoted to an open-source command-line A* parking
simulation \cite{autopark} as a lean implementation baseline. This approach was
well-suited to the team's existing strengths: the repository was cloneable and
runnable locally using standard tools (Git, Python, and the command line), and
the codebase was accessible enough to modify without requiring deep familiarity
with a large simulation framework. This allowed us to direct our effort toward
the core technical problem of extending single-body path planning to an
articulated tractor-trailer system, rather than simulator configuration.

\subsection{Trailer Extension and Kinematic Modeling}

The core implementation contribution of this project is the modification of the
open-source baseline to incorporate trailer dynamics. In the original codebase,
path planning operates over a single rigid vehicle body. We extended this to a
two-body model by adding a trailer unit connected to the tractor at a hitch point,
with the trailer's heading governed by the following kinematic relationship:

\begin{equation}
\dot{\psi}_t = \frac{v}{L_t} \sin(\psi - \psi_t)
\end{equation}

where $\psi$ is the tractor heading, $\psi_t$ is the trailer heading, $v$ is the
vehicle speed, and $L_t$ is the trailer length. This formulation captures the
inverse steering response of the trailer during reversing maneuvers and enforces
the geometric coupling between the two bodies throughout path execution.

\begin{figure}[htbp]
\centering
\includegraphics[width=\columnwidth]{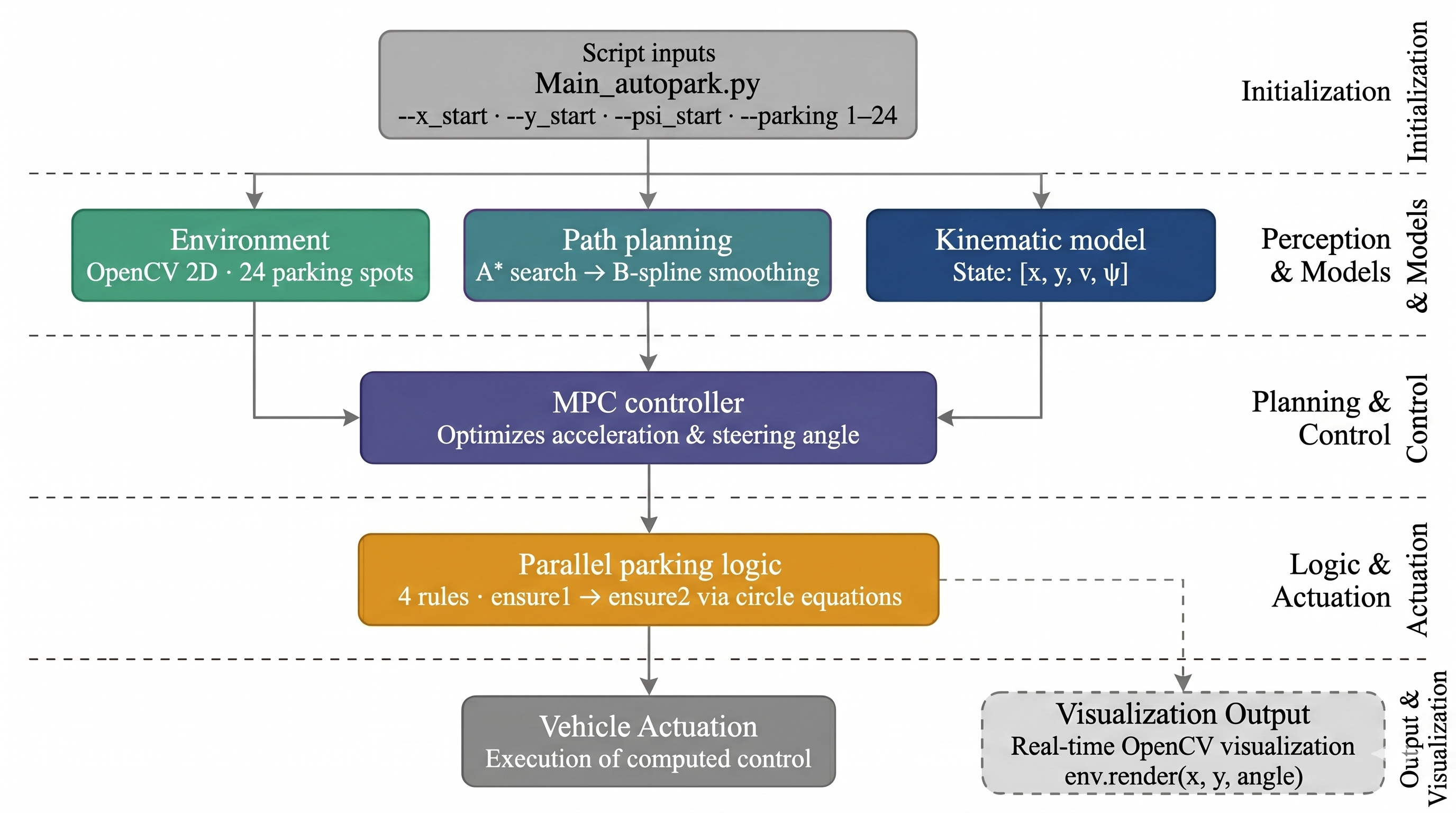}
\caption{Architecture of the adapted open-source simulation baseline, showing
the pipeline from script inputs through path planning, kinematic modeling,
and MPC control to vehicle actuation and visualization output.}
\label{fig:code_arch}
\end{figure}

Figure~\ref{fig:code_arch} illustrates the full pipeline of the adapted simulation,
from script initialization through to vehicle actuation and real-time visualization.

\subsection{Results}

Table~\ref{tab:results} summarizes the outcomes of our implementation effort
across the environments and configurations we evaluated, including basic high-level metrics collected from the simulation run.

\begin{table}[htbp]
\caption{Implementation Outcome Summary}
\label{tab:results}
\centering
\begin{tabular}{>{\raggedright\arraybackslash}p{2.8cm}
                >{\raggedright\arraybackslash}p{1.4cm}
                >{\raggedright\arraybackslash}p{3.1cm}}
\toprule
\textbf{Component / Task} & \textbf{Status} & \textbf{Notes} \\
\midrule
CARLA simulation setup      & Not adopted & Hardware access and scheduling constraints \\
Gazebo / MATLAB evaluation  & Not adopted & Unfamiliar tooling; lab access limitations \\
A* baseline integration     & Complete    & Simulation runs in command-line environment \\
Trailer kinematic model     & Complete    & Tractor-trailer two-body model implemented \\
Path planning (fixed route) & Successful  & Vehicle and trailer navigate target path \\
Dynamic / arbitrary paths   & In progress & Current paths are fixed; generalization is future work \\
Full sensor fusion pipeline & Planned     & Architecture designed; not yet implemented \\
\bottomrule
\end{tabular}
\end{table}

The adapted simulation successfully plans and executes a parking path for a
tractor-trailer vehicle, correctly modeling the motion of both the tractor and
the trailer throughout the maneuver. The trailer heading updates in real time
as the tractor navigates the path. Path computation completes within a short
planning phase before vehicle motion begins, after which the MPC controller
executes the trajectory in real time. During testing, jackknifing behavior was
observed in certain configurations, indicating that while the trailer kinematic
model is functional, hitch angle constraint enforcement is not yet robust across
all maneuver conditions. This is consistent with the current state of the
implementation as a proof-of-concept rather than a production system, and
addressing jackknife prevention is identified as a priority for future work.
The primary limitation beyond jackknifing is path generalizability: the simulation
produces a consistent result for the fixed tested route, and work remains to
support arbitrary start-goal configurations and dynamic obstacle avoidance.

\begin{figure}[htbp]
\centering
\includegraphics[width=\columnwidth]{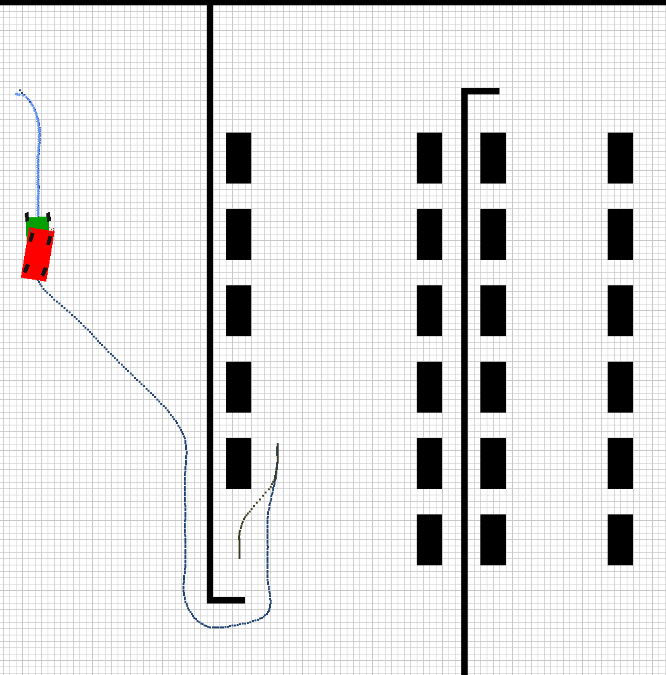}
\caption{Command-line simulation showing the tractor-trailer vehicle navigating
toward the target parking space. The trailer unit is rendered behind the tractor
and its heading updates dynamically throughout the maneuver.}
\label{fig:sim_start}
\end{figure}

\begin{figure}[htbp]
\centering
\includegraphics[width=\columnwidth]{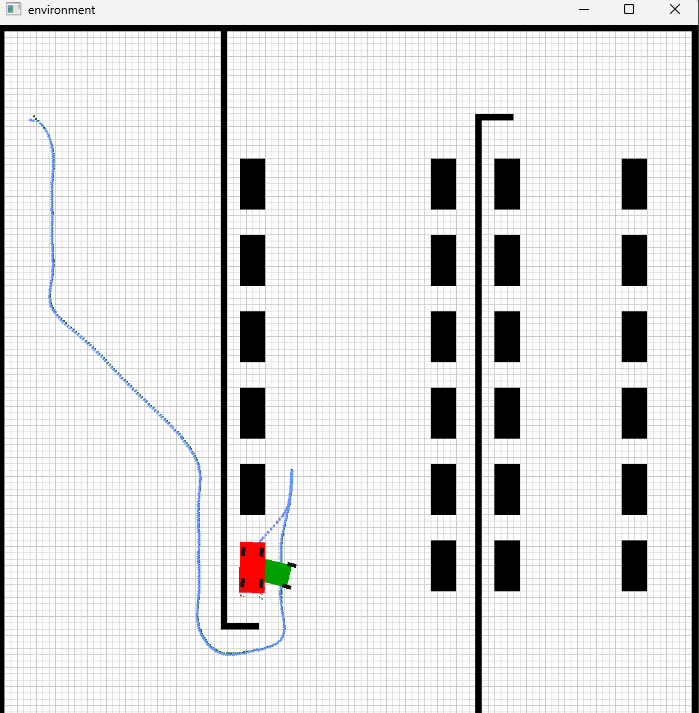}
\caption{Simulation output after completion of the parking maneuver, showing
the tractor-trailer vehicle aligned with the target parking space. The trailer
heading has not yet fully converged to match the tractor orientation in our implementation.}
\label{fig:sim_end}
\end{figure}


\section{Discussion and Challenges}
\label{sec:discussion}

Autonomous truck parking is a multi-layered problem that cannot be solved by any
single technique in isolation. The framework presented in this paper demonstrates
that effective solutions require tight integration across perception, planning,
control, and infrastructure awareness. Each layer depends on the others: accurate
perception enables reliable planning, feasible plans enable safe control, and
infrastructure data enables the system to identify actionable parking targets in
the first place.

Our implementation experience reinforced this dependency. The decision to pivot
away from high-fidelity simulators was ultimately driven by the realization that
the vehicle kinematic model, specifically the articulated tractor-trailer coupling,
was the foundational challenge that had to be addressed before any higher-level
system integration would be meaningful. Getting the kinematics right in a simpler
environment first provides a validated model that can later be ported into a more
capable simulation platform.

The incorporation of shared mapping in the broader proposed architecture introduces
an important collaborative dimension, allowing the system to benefit from the
collective experience of a fleet rather than relying solely on onboard sensing.
This is particularly valuable for truck parking, where infrastructure conditions
are highly variable and difficult to capture in static maps.

Several significant challenges arose during this project and remain in any effort
to realize a fully autonomous truck parking system:

\begin{itemize}

\item \textbf{Simulation environment accessibility:} High-fidelity simulators
such as CARLA posed practical barriers including large installation size and the
need for dedicated lab hardware, while Gazebo and MATLAB required domain expertise
the team did not have in visual vehicle simulation. These constraints collectively
made iterative development in those environments infeasible within the project
timeline, motivating the pivot to a locally-runnable open-source baseline.

\item \textbf{Computational requirements:} Real-time NMPC and transformer-based
fusion impose substantial processing demands that must be met within the latency
constraints of a live control loop. These demands were not addressed in the current
proof-of-concept but will be central in any production system.

\item \textbf{Limited truck-specific datasets:} Most publicly available autonomous
driving datasets are collected from passenger vehicles, limiting the availability
of training and evaluation data specific to articulated truck dynamics.

\item \textbf{Complex articulated dynamics:} The nonlinear coupling between tractor
and trailer motion, particularly during reversing, makes trajectory planning and
control significantly more difficult than for single-body vehicles. This was the
central technical challenge of our implementation work.

\item \textbf{Path generalizability:} The current simulation produces consistent
results for a fixed route but does not yet generalize to arbitrary parking targets.
Extending A* search over the full configuration space of the articulated vehicle
is required to address this.

\item \textbf{Hitch angle constraint enforcement:} While the trailer kinematic
model correctly captures articulated vehicle dynamics, jackknifing was observed
during certain tested configurations. Robust enforcement of the hitch angle
safety constraint across arbitrary maneuvers remains an open problem for future
implementation work.

\end{itemize}


\section{Conclusion}
\label{sec:conclusion}

This paper presented a research synthesis and proof-of-concept implementation
for autonomous truck parking, addressing one of the most mechanically demanding
problems in autonomous vehicle systems. Our primary contributions are threefold.
First, we conducted a systematic survey of the algorithmic components required
for a full autonomous truck parking pipeline, including sensor fusion, localization,
Hybrid A* motion planning, NMPC trajectory refinement, LQR control, and
data-driven parking infrastructure, and synthesized them into a proposed unified
system architecture. Second, we evaluated multiple simulation environments (CARLA,
Gazebo, MATLAB) for truck-specific applicability and identified the critical gap
in articulated vehicle support that limits their use for this problem. Third, we
adapted an open-source A* path planning simulation to incorporate a tractor-trailer
kinematic model, producing a working proof-of-concept that models trailer motion
throughout a parking maneuver, though robust jackknife prevention across all
configurations remains future work.

The key finding of our implementation work is that the articulated kinematic model
is the irreducible foundation of any truck parking system: until the two-body
dynamics are correctly represented, higher-level planning and control layers cannot
meaningfully be integrated or evaluated. Our adapted simulation validates this
model in a controlled environment and provides a baseline for future development.

Future work will focus on three areas. First, extending the path planner to support
arbitrary start-to-goal configurations rather than fixed routes, enabling
generalized parking in varied environments. Second, integrating the validated
kinematic model into a more capable simulation platform such as Gazebo or MATLAB
for higher-fidelity testing. Third, incrementally adding the perception, sensor
fusion, and infrastructure layers described in the proposed architecture, working
toward a full end-to-end implementation.

\section*{Acknowledgment}

The authors would like to thank Florida Gulf Coast University
for its support.



\appendix
\section{Individual Contributions}

The work described in this paper reflects the following team member contributions.

\textbf{George Alenchery:} Conducted research on autonomous parking systems for
articulated vehicles, investigated the Hybrid A* algorithm as the proposed planning
method, and identified the open-source A* simulation GitHub repository used as the
implementation baseline \cite{autopark}.

\textbf{Thomas Jeske:} Researched and reported on current limitations the trucking
sector faces and proposed solutions regarding parking. Investigated the usability
of the Viper Lab. Assisted in the discovery of a key GitHub repository that helped
show what was possible for the project's implementation, and suggested the trailer
edit for a final output closer to the project goals.

\textbf{Lentz Fortune:} Added an inheritance class called ``Trailer Dynamics'' and
edited the Environment.py file in the Automatic Parking GitHub repository in order
to render a trailer to the back of the car parking simulation, allowing the
repository to better reflect the project goals.

\textbf{Jordan Fletcher:} Researched sensor fusion and how systems like TransFuser
combine LiDAR and camera data to improve environmental understanding in autonomous
vehicles. Helped with writing the paper and kept progress updates organized.

\textbf{Amber Jones:} Researched real-time object detection techniques using
LiDAR-camera fusion integrated with adaptive dynamic programming for autonomous
truck-trailer driving, contributing to the understanding of how perception and
control strategies are combined.

\textbf{Tristan Lindo-Slones:} Researched two-layered architecture for autonomous
parking, including deep reinforcement learning combined with a kinematic model for
accurate alignment. This research informed the approach to managing complex
articulated surroundings during reversing and docking.

\textbf{Tova Quinones:} Contributed to the research and conceptual design of the
system by investigating motion planning and sensor fusion, focusing on trajectory
planning, control systems, and multi-source data fusion in limited parking
environments.

\end{document}